\documentclass{article}

\usepackage{arxiv}
\usepackage{siunitx}
\usepackage[utf8]{inputenc} 
\usepackage[T1]{fontenc}    
\usepackage{hyperref}       
\usepackage{url}            
\usepackage{booktabs}       
\usepackage{amsfonts}       
\usepackage{nicefrac}       
\usepackage{microtype}      
\usepackage{lipsum}
\usepackage{graphicx}
\graphicspath{ {./images/} }
\usepackage{amsmath}

\title{Distil the informative essence of loop detector data set: Is network-level traffic forecasting hungry for more data?}

\author{
 Guopeng Li\\
  Civil Engineering and Geosciences\\
  Delft University of Technology, Netherlands\\
  Email: g.li-5@tudelft.nl\\ \\
   \And
 Victor L. Knoop\\
  Civil Engineering and Geosciences\\
  Delft University of Technology, Netherlands \\
  \And
 \textbf{J.W.C. (Hans) van Lint}\\
  Civil Engineering and Geosciences\\
  Delft University of Technology, Netherlands\\
}

\begin{document}
\maketitle
\begin{abstract}
Network-level traffic condition forecasting has been intensively studied for decades. Although prediction accuracy has been continuously improved with emerging deep learning models and ever-expanding traffic data, traffic forecasting still faces many challenges in practice. These challenges include the robustness of data-driven models, the inherent unpredictability of traffic dynamics, and whether further improvement of traffic forecasting requires more sensor data. 
In this paper, we focus on this latter question and particularly on data from loop detectors. To answer it, we propose an uncertainty-aware traffic forecasting framework to explore how many samples of loop data are truly effective for training forecasting models. Firstly, the model design combines traffic flow theory with graph neural networks, ensuring the robustness of prediction and uncertainty quantification. Secondly, evidential learning is employed to quantify different sources of uncertainty in a single pass. The estimated uncertainty is used to "distil" the essence of the dataset that sufficiently covers the information content. 
Results from a case study of a highway network around Amsterdam show that, from 2018 to 2021, more than 80\% of the data during daytime can be removed. The remaining 20\% samples have equal prediction power for training models. This result suggests that indeed large traffic datasets can be subdivided into significantly smaller but equally informative datasets. From these findings, we conclude that the proposed methodology proves valuable in evaluating large traffic datasets' true information content. Further extensions, such as extracting smaller, spatially non-redundant datasets, are possible with this method.
\end{abstract}

\keywords{Traffic forecasting \and Data distillation \and Uncertainty quantification}

\section{Introduction}

Traffic forecasting is an important element of Intelligent Transportation Systems (ITS). Reliable traffic state prediction enables transportation planners and road authorities to anticipate future traffic congestion and guide traffic flow for safer and more efficient travels \cite{boukerche2020machine}. In the past several decades, macroscopic traffic state forecasting has been intensively studied. A wide range of models have been proposed in the literature. Yuan et al. \cite{yuan2021survey} provide a comprehensive review. Recently, due to the growing availability of traffic state data and the advancements in data-driven techniques, particularly deep learning, there has been a notable trend towards utilizing deep neural networks (DNNs) for predicting network-level traffic conditions. Akhar et al. \cite{akhtar2021review} thoroughly review the application of artificial intelligence (AI) methods in traffic congestion prediction. Moreover, Jiang et al.\cite{jiang2022graph} focus specifically on surveying the use of graph neural networks for traffic forecasting. For more comprehensive information, we refer the readers to these review papers.

Although deep learning and other data-driven models are demonstrated to be accurate in many traffic forecasting tasks, improving the prediction accuracy is also becoming more and more difficult. We observe that even using increasingly larger loop detector data sets and more sophisticated neural networks, the improvement in prediction accuracy remains marginal. PEMS-BAY and METR-LA, released by Li et al. \cite{li2017diffusion}, are two examples of popular open traffic state data sets used for comparing model performances. Since the DCRNN model proposed by Li et al. \cite{li2017diffusion} in 2018, the mean absolute error (MAE) on METR-LA has been improved by 0.32 mph over the past five years, with Traffic-Transformer \cite{cai2020traffic} being the current state-of-the-art. Similarly, for PEMS-BAY, the MAE has been improved by 0.30 mph only. Such insignificant improvements hold little relevance for prediction-related tasks like traffic control. Several research papers have attempted to shed light on this limited improvement from different perspectives. For instance, Manibardo et al. \cite{manibardo2021deep} argue that deep learning models do not consistently exhibit superior performance in many case studies. Li et al. \cite{li2022estimate} interpret the limited predictability from the standpoint of uncertainty, suggesting that the high uncertainty of traffic dynamics, stemming from limited observability, hampers significant accuracy improvement. All these lead to a crucial underlying research question: 

\emph{Do loop detector data sets provide sufficient and diverse information to support training large and complex data-driven prediction models?} 

Addressing this question can provide valuable insights into the future research direction of this field.

\subsection{Data vs. modelling for traffic forecasting}

Data and modelling are central topics in traffic forecasting. Despite many new data sources becoming available, many papers still use aggregated traffic flow, speed, and density data collected by loop detectors to predict future traffic conditions \cite{yuan2021survey}. The focus in this large body of work has primarily been on adapting modelling techniques in the machine learning domain, particularly various deep neural network structures, to address the traffic forecasting problem. Whereas much effort and attention are dedicated to modelling and accuracy benchmarking, limited attention is given to how informative the used (loop detector) data are for the prediction task at hand. A widely held intuition is that the relationship between data and modelling follows an upward spiral, that is, as the size of available data increases, researchers explore different model structures to effectively leverage the information provided by the data until the model's performance approaches its limit \cite{sun2017revisiting} and attention shifts back to data set expansion, and so forth. This paradigm has been successful in numerous domains, particularly in the fields of computer vision (CV) \cite{o2020deep} and natural language processing (NLP) \cite{torfi2020natural}.

However, network-level traffic forecasting radically differs regarding data collection. First, traffic systems involve many hidden and intertwined variables that govern the underlying traffic dynamics. The observed traffic state is just the manifestation of these factors. The information comprised in the loop detector data typically cannot support predicting longer-term traffic states, which heavily depend on unknown factors such as traffic demand and stochastic driving behaviours \cite{tampere2005behavioural}. Moreover, each city and each road network has its unique demand pattern and topology, making data transfer between different road networks challenging \cite{krishnakumari2018understanding}.  
Finally, it is not a priory clear that expanding macroscopic traffic data sets also implies expanding the information content. For example, a typical workday can provide congestion patterns of two peak hours only and there are uncongested holidays throughout the year. Another concern is that traffic congestion often exhibits strong periodicity and regularity. For example, Lopez et al. \cite{lopez2017revealing} demonstrate the regular patterns of highway congestion in the Netherlands. Nguyen et al. \cite{nguyen2019feature} show that corridor-level congestion patterns can also be clustered into 4 major classes. This regularity further diminishes the informative aspect of a loop detector data set. For instance, if a road is congested during the evening peak hour at the same time every day, collecting data repeatedly becomes redundant and unnecessary.

In summary, whereas improving traffic forecasting models holds significant importance and remains a key area of research, it is equally important to assess how much information content loop detector datasets contain. In this study, we aim to address this issue by focusing on uncertainty quantification and reducing dataset size as our primary perspectives.

\subsection{Uncertainty quantification and data set distillation}

To evaluate how informative a data set is one must consider two aspects of uncertainty \cite{abdar2021review}. The first one is the inherent unpredictable component (randomness) of the dynamic process, like the variance of a Gaussian process, caused by limited types of data, the so-called \emph{data uncertainty}. The second one is the diversity of the comprised samples. A basic assumption from information theory is that rare congestion patterns may provide more information than common and recurring patterns. The quantity of ``entropy'' measures this rarity and the corresponding amount of information. This aspect is known as \emph{knowledge uncertainty}. Data uncertainty reflects the information contained within specific types of loop detector data. Data uncertainty is often utilized to establish a lower bound for prediction accuracy \cite{li2022estimate}. On the other hand, knowledge uncertainty measures the effective amount of data present in a given data set, which can be used to ``distil'' a large data set down to its essential and informative component \cite{cazenavette2022dataset}.

The uncertainty-based data set distillation method aims to retain rare and informative samples with high knowledge uncertainty while filtering out common, repetitive samples. For historical data sets, the samples are ranked based on estimated rarity (knowledge uncertainty) from low to high. Then, a threshold is sought that allows removing samples below this value without adversely affecting the training of well-formed predictors. If the number of samples above this threshold (informative samples) is significantly lower than the total number, the data set is considered uninformative, even if its size is large. When deploying the model in a live data stream, valuable samples can also be identified and selected using knowledge uncertainty. This approach allows us to investigate whether training network-level traffic forecasting models truly benefits from bigger data in practical scenarios.

\subsection{Contributions and outline}
In this study, we propose a unified framework to provide accurate traffic state predictions while also offering uncertainty estimation. To address the primary research question of whether loop detector data sets are informative enough, we conducted a case study on a highway network around Amsterdam. The key contributions of this research can be summarized as follows:
\begin{itemize}
    \item Identify informative samples from the data set and data stream by an uncertainty-aware traffic forecasting framework.
    \item Illustrate that only a small essential part of loop detector data is indispensable for training traffic forecasting models.
    \item Conclude that traffic forecasting requires more diverse data types instead of larger loop detector data.
\end{itemize}

This paper is structured as follows. The proposed method is first explained in the methodology section. Next, the used model structure is presented. Then, the experiments on a highway network are carried out and the results are shown and discussed. Finally, the last section draws conclusions and proposes several relevant research topics.

\section{Method}
\label{sec: theory}

This section presents the method used in this study. Model design, uncertainty quantification, and data set distillation will be introduced sequentially.

\subsection{Model design principles}

Model design is crucial for robust prediction and uncertainty estimation. Data-driven models, particularly deep learning models, are sensitive to correlated features. However, correlation alone does not necessarily imply causal relationships. For instance, consider two distant roads that are congested simultaneously during evening peak hours. Although their traffic conditions are highly correlated, there is no physical connection between them. Using the traffic state of one road to predict the other one may improve accuracy in most cases, but the prediction will fail when their traffic states de-correlate. This phenomenon is the so-called \emph{causal confusion} \cite{de2019causal} and thus the estimated uncertainty can be inflated or deflated. Therefore, using a robust model that indeed learns the expected associations is the prerequisite for properly addressing the primary research question. In this study, the model design involves two parts: the "causation-like" propagation of traffic states and the incorporation of additional fluctuations.

To predict a traffic quantity, denoted as $X$ (e.g., traffic flow), the fundamental concept is that its value at location $i$ in the near future, $X_i^{t+1}$, can be determined as a linear combination of the (functions of) traffic states of surrounding locations (including the target location $i$ itself), denoted as $\mathcal{N}(i, d)$ ($d$ is the radius of the neighbour area), along with an additional fluctuation term $r$:
\begin{equation}
    X_i^{t+1} = \sum_{j \in \mathcal{N}(i, d)} w_{ji}^t f(X_j^t) + r_i^t
    \label{eq: basic}
\end{equation}

In this context, the coefficient $w_{ji}$ quantifies the impact of location $j$ on the target location $i$. The term $r_i^t$ accounts for external factors like driving behaviours or varying traffic demand. Without the fluctuation term, the weighted average alone cannot capture abrupt transitions in traffic conditions, such as the onset of new congestion or the dissipation of existing congestion. It's important to note that both $w_{ji}^t$ and $r_i^t$ are not constant. Their values depend on the traffic states within $\mathcal{N}(i, d)$:
\begin{equation}
    \begin{aligned}
        & w_{ji}^t = w_{ji}(X^t_{\mathcal{N}(i, d)}, c)\\
        & r_i^t = r_i(X^t_{\mathcal{N}(i, d)})
    \end{aligned}
\end{equation}

Here we add a parameter $c$ to indicate the location $j$ is upstream or downstream of $i$. Traffic flow is directional in nature so the impact is also directional. To stabilize the prediction, it is generally presumed that the sum of weights equals 1 and the fluctuation term is bounded:
\begin{equation}
    \begin{aligned}
        &\sum_{j \in \mathcal{N}(i, d)} w_{ji}^t = 1\\
        &r_i^t \in [-R_i, R_i],\ \ R_i>0
    \end{aligned}
    \label{eq: assumptions}
\end{equation}

The sign and value of $r_i^t$ depend on the location (infrastructure) and the current traffic state. The boundary $R_i$ can be approximated by the maximum charge/discharge rate of highways (around 1800 veh/lane/h) in the case of traffic flow prediction and the speed limit for speed prediction.

The description above can be easily integrated into data-driven approaches, such as the popular graph attention networks \cite{velickovic2017graph}. Nevertheless, in order to ensure that the learned $w_{ji}^t$ correctly represents the causal contributions, incorporating traffic flow theory into the modelling design is necessary. For one-dimensional traffic flow (on a uniform corridor), a fundamental description is the LWR model \cite{lighthill1955kinematic, richards1956shock}. If the density is denoted as $\rho$, it says:
\begin{equation}
    \frac{\partial \rho(x,t)}{\partial t} + c_q(\rho) \frac{\partial \rho(x,t)}{\partial x} = s(x,t)
    \label{eq: basic lwr}
\end{equation}

Eq.\eqref{eq: basic lwr} is an advection function. If the source term $s(x,t) = 0$ and the speed rate function $c_q(\rho)$ satisfies the following inequality:
\begin{equation}
    \frac{d c_q(\rho(x, t=0))}{dx} \geq 0
\end{equation}
then a unique classical solution exists. The traffic state keeps the same along the characteristic curves, shown as running forward traffic stream or back-propagating stop-and-go shock waves on the spatiotemporal graphs:
\begin{equation}
    \frac{dx}{dt} = c_q(\rho)
\end{equation}

The values on different characteristic curves can be close (such as the adjacent stop-and-go waves) so $\mathcal{N}(i, d)$ cannot include two characteristic curves that have the same value when performing the prediction. Otherwise, the data-driven model may be confused with the true causal relationship (Which shock wave leads to the predicted congestion?). Therefore, when designing a data-driven predictor, the following condition must be met, with the lower bound being the CFL condition \cite{courant1967partial}:
\begin{equation}
    \Delta x \in [c_q(\rho)\Delta t, 2c_q(\rho)\Delta t)
    \label{eq: range boundary}
\end{equation}

Here $\Delta t$ is the time interval. With this constraint, the learnt $w_{ji}^t$ is expected to correctly reflect the causal influence. For network-level traffic forecasting, although there are no characteristic curves for the advection equation on a directed graph, the same principle holds. Because the underlying speed limit of information propagation is the same. 

In summary, the rules of model design are as follows:

\begin{itemize}
    \item Both the impact of the surrounding locations and the fluctuation term must be learnt from the dynamic traffic state within an area.
    \item The impact must consider the relative positions of the two locations, especially which one is upstream.
    \item For each step prediction, the model should only aggregate features in a specific range from the last observations. The range is determined by the propagation speed of information and the time interval.
\end{itemize}

By applying the principles above, a data-driven model can learn explainable causal spatiotemporal relationships. Note that if the prediction contains two quantities (e.g., speed and flow), their predictions must be interpreted separately.

\subsection{Single-pass uncertainty quantification}

After designing a proper model structure, robust prediction and reasonable uncertainty quantification become possible. Considering the amount of historical data to handle, in this study, we employ a single-pass uncertainty quantification method proposed by Amini et al.\cite{amini2020deep} and also used in regression prior networks \cite{malinin2020regression}, the so-called \emph{deep evidential learning} (DEL). Compared with Monte-Carlo dropout \cite{gal2016dropout}, deep ensembles \cite{lakshminarayanan2017simple}, and Bayesian neural networks \cite{blundell2015weight}, DEL has significantly lower spatial complexity and higher inference speed because it is not necessary to run multiple models in parallel or run one model multiple time for variational inference. In this subsection, we will briefly introduce the DEL method.

The basic idea of DEL is the same as other methods. Distinguishing the data vs. knowledge uncertainty requires a higher order of distribution, e.g. the distribution of the parameters of the predicted distribution. Assume that the predicted traffic quantity (such as speed, denoted as $x$) obeys a Gaussian prior $x \sim \mathcal{N}(\mu, \sigma^2)$ with unknown mean and variance. This unknown variance itself represents the data uncertainty (irreducible randomness). However, the mean and the variance need to be learnt from a set of observed point values denoted as $\{\Tilde{x}\}$. Knowledge uncertainty represents how confident we are about the learned mean and variance. High knowledge uncertainty means that not enough samples are observed in the training set to accurately estimate the parameters of the Gaussian prior. Therefore, we also regard $\mu$ and $\sigma^2$ as two random variables. If their probability distribution is concentrated, knowledge uncertainty is low. For algebraic convenience, DEL assumes that $\mu$ and $\sigma^2$ obey the \emph{conjuagate prior} of Gaussian. $\mu$ is Gaussian and $\sigma^2$ obeys an \emph{inverse-gamma distribution}:
\begin{equation}
    \begin{aligned}
        &\mu \sim \mathcal{N} (\lambda, \sigma^2/\nu)\\
        &\sigma^2 \sim \Gamma^{-1} (\alpha, \beta)
    \end{aligned}
\end{equation}

Therefore, the joint distribution of $(\mu, \sigma^2)$ obeys an \emph{Normal-Inverse-Gamma} (NIG) distribution:
\begin{equation}
    p(\mu, \sigma^2 | \lambda, \nu, \alpha, \beta) = \frac{\beta^{\alpha}\sqrt{\nu}}{\Gamma(\alpha) \sqrt{2\pi\sigma^2}} (\frac{1}{\sigma^2})^{\alpha + 1}\times \exp{\{-\frac{2\beta + \nu(\lambda-\mu)^2}{2\sigma^2}\}}
\end{equation}

By marginalizing out $\mu$ and $\sigma$, the posterior of $x$ obeys a Student's t-distribution:
\begin{equation}
    p(x|\lambda, \nu, \alpha, \beta) = lst(x; \lambda, \frac{\beta(1+\nu)}{\nu\alpha}, 2\alpha)
    \label{eq: student}
\end{equation}
whose variance (and also the total uncertainty) is:
\begin{equation}
    \text{Var}(x) = \frac{\beta(\nu+1)}{\nu(\alpha-1)}
    \label{eq: total var}
\end{equation}

According to the total law of variance, we have:
\begin{equation}
    \text{Var}(x) = \text{E}(\sigma^2) + \text{Var}(\mu) = \frac{\beta}{(\alpha-1)} + \frac{\beta}{\nu(\alpha-1)}
    \label{eq: decompose}
\end{equation}

The detailed derivative above is provided in Amini et al. \cite{amini2020deep}. The first term in Eq.\ref{eq: decompose} is the mean of estimated variance, which represents data uncertainty. The second term is the variance of the estimated mean values, which represents knowledge uncertainty. This formula can be intuitively interpreted by the so-called \emph{t-test} \cite{kim2015t}. When Eq.\ref{eq: student} are estimated by likelihood maximization, we are in fact assimilating the mean, variance, and also the \emph{degree of freedom} $2\alpha$, which represents the ``virtual'' number of observations in the t-test. When $\alpha \rightarrow +\infty$ (enough observations), then Eq.\ref{eq: student} converges to a Gaussian (the distributions of $\mu$ and $\sigma^2$ all converge to a single value or the so-called Dirac function). A higher $\alpha$ value means that the prior Gaussian can be estimated accurately, and thus low knowledge uncertainty (common cases). Conversely, lower $\alpha$ means fewer observations so knowledge uncertainty is high (in rare cases).

The negative log-likelihood (NLL) of Student's t distribution to be minimized is:
\begin{equation}
    \mathcal{L}_{s} = \frac{1}{2}\log{(\pi/\nu)} - \alpha \log{[2\beta(\nu+1)]}
    + (\alpha + \frac{1}{2}) \log{[(\Tilde{x}_i - \lambda)^2\nu + 2\beta(1+\nu)]}
    + \log{[\frac{\Gamma(\alpha)}{\Gamma(\alpha + 0.5)}]}
\end{equation}

Learning the Student's t distribution from data can determine $\lambda$ and $\alpha$, but $\beta$ and $\nu$ cannot be learnt directly because $\beta$ determines the distribution of $\sigma^2$ only. However, during training, $\sigma^2$ for each input is unknown (we only have the ground truth of output, an observation). Therefore, additional regularization must be induced. Here we use the trade-off between the estimated standard deviation and the observed MAE. The regularization formula is:
\begin{equation}
    \mathcal{L}_{r} = [\ \underbrace{|\Tilde{x}_i - \lambda|\ /\ (C\times \sqrt{\frac{\beta}{\alpha-1}})}_\text{observed error/estimated error} - 1\ ]* (\nu + \alpha)
    \label{eq: reg}
\end{equation}
where $C$ is a normalization constant derived from the Gauss error function \cite{gautschi1970efficient}:
\begin{equation}
    C = \sqrt{2}\times \text{erf}^{-1}(1/2) \approx 0.674
\end{equation}

Eq.\ref{eq: reg} is a novel regularization method that is different from the original MAE regularization used in DEL \cite{amini2020deep}. It can be interpreted as follows. When the standard deviation of the ground truth to the predicted mean value conforms well with the estimated mean-average-deviation of the Gaussian prior (the first term in the parenthesis tends to 0), then this is a low knowledge uncertainty case (enough observations are given, we can accurately estimate the Gaussian prior), and thus $\nu$ is large. Otherwise, if the deviation is large, then the input is a rare case and $\nu$ should be small. With this regularization, $\beta$ and $\nu$ can be determined by minimizing the log-likelihood. It is useful to note that using the ratio of the observed MSE and the estimated data uncertainty variance theoretically also works, but the non-linear scale (square) makes the training process less stable than the used one. The goal is to minimize the following overall objective function:
\begin{equation}
    \mathcal{L}_{t} = \mathcal{L}_{s} + \epsilon  \mathcal{L}_{r}
    \label{eq: loss}
\end{equation}
where $\epsilon$ is a hyper-parameter that adjusts the level of knowledge uncertainty. During the inference, the model gives 4 parameters in Eq.\eqref{eq: student} for each point so both types of uncertainty can be quantified by Eq.\eqref{eq: decompose}. For network-level traffic forecasting, we can use 4 vectors to denote the 4 parameters at all locations at a specific time step $t$:
\begin{equation}
    [\lambda, \nu, \alpha, \beta] \Rightarrow [\pmb{M}^t, \pmb{V}^t, \pmb{A}^t, \pmb{B}^t]
\end{equation}

Next, we will introduce how to use the estimated knowledge uncertainty to distil the training set and to collect valuable samples in the online data stream.

\subsection{data set distillation}

As discussed in the introductory section, data uncertainty estimates the limit of prediction errors and knowledge uncertainty quantifies how rare each sample is. Now we "distil" the traffic data and only preserve the high-value data. The procedure is:
\begin{itemize}
    \item[1] Training an uncertainty-aware traffic forecasting model on the training set. Then rank all samples from the lowest to the highest by knowledge uncertainty.
    \item[2] Find a critical value of knowledge uncertainty such that removing all samples lower than it does not negatively impact the model's accuracy.
    \item[3] When deploying the model in the live data stream, only preserve those congestion patterns whose knowledge uncertainty is higher than the threshold.
\end{itemize}
In this way, we can explore how much data in the historical data set is indeed necessary for training data-driven forecasting models. Next, we will introduce the proposed model structure.

\section{Model}\label{sec: model}
To fulfil the requirements discussed in the model design section, we inherit and extend the graph attention mechanism proposed in our previous study, the so-called \emph{dynamic graph convolution (DGC) module} \cite{li2021multistep}. DGC module generates state-dependent dynamic kernels and then applies graph convolution to give predictions. The original DGC is only used for speed prediction and the fluctuation term is not dynamic. In this study, several key modifications are made for implementing uncertainty quantification and multiple traffic quantities prediction.

The first step is formulating the desired output and the relevant input quantities. Traffic flow theory tells that at least two quantities amongst speed, flow, and density should be given to describe the traffic state. In this study, we choose speed and flow. For uncertainty quantification, considering that speed is a better congestion indicator than flow (low traffic flow can be either free-flowing or congested), we only quantify the uncertainty in predicted speed. The backbone of the proposed model is an RNN encoder-decoder \cite{cho2014learning}. The encoder extracts the features from the recent multistep observations and passes the encoded hidden state to the decoder. The decoder then unrolls the network-level predictions step by step. Both the input and the output of one RNN cell must include both the prediction of the traffic state and the 3 additional uncertainty parameters:
\begin{equation}
    [\pmb{M}_v^t, \pmb{M}_q^t, \pmb{V}^t, \pmb{A}^t, \pmb{B}^t]
\end{equation}

\begin{figure}[h!]
\centering
\includegraphics[width=0.65\linewidth]{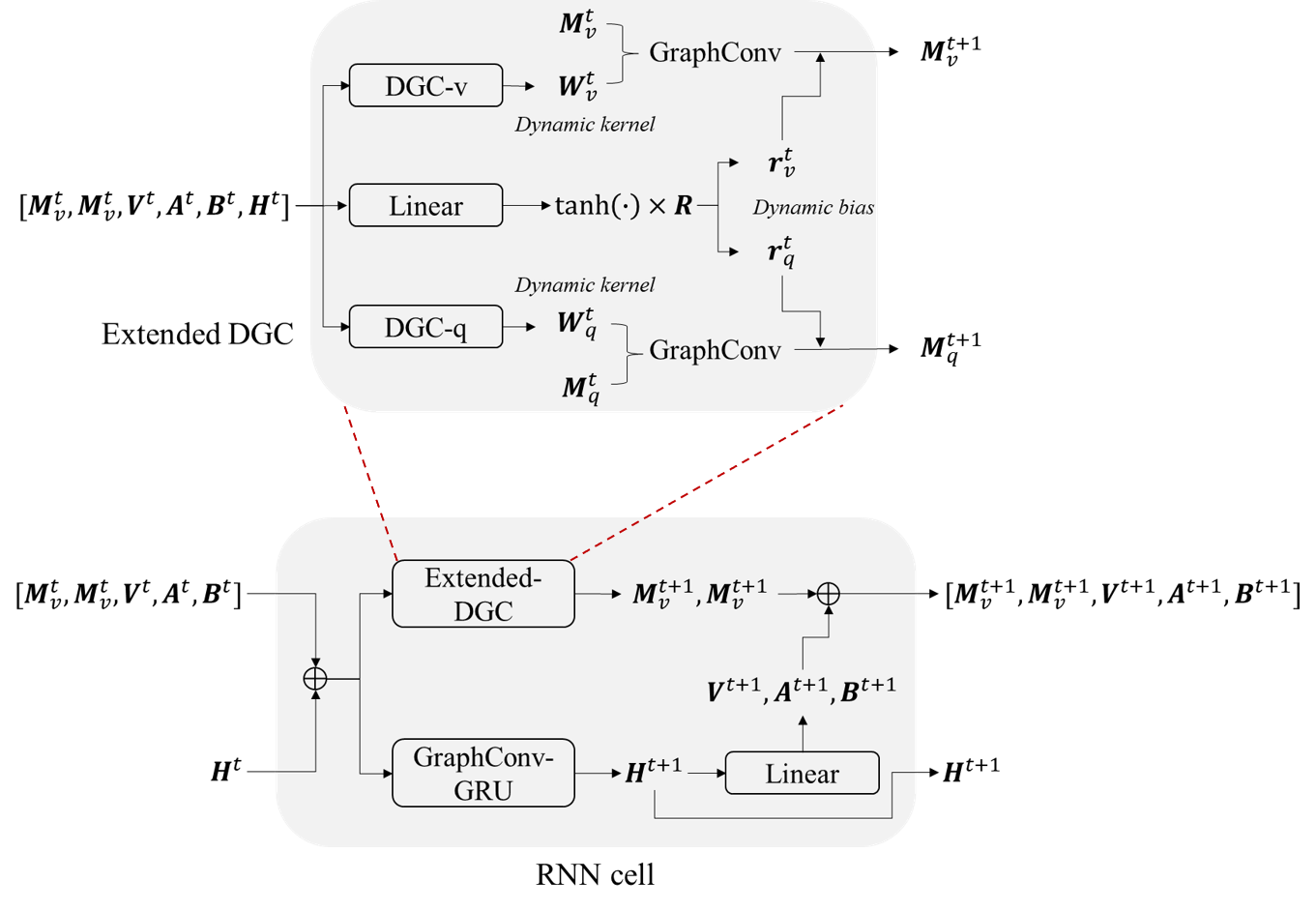}
\caption{The selected highway network. We only consider the clockwise driving directions in this study.} \label{fig: modelarchi}
\end{figure}

Here $\pmb{M}_v^t, \pmb{M}_q^t$ represent the mean values of speed and flow respectively. $\pmb{V}^t, \pmb{A}^t, \pmb{B}^t$ are the uncertainty parameters for speed prediction. The structure of one RNN cell is shown in Fig.\ref{fig: modelarchi}. The input is first concatenated with the hidden state $\pmb{H}^t$. Next, an extended DGC module generates the mean value of the speed and flow for the next time step. In the other branch, a graph convolution GRU cell gives the new hidden state $\pmb{H}^{t+1}$. In this cell, the gates in a normal GRU cell \cite{cho2014learning} are replaced by static graph convolutional layers \cite{chen2020simple}. The other uncertainty parameters for the next step are generated from the new hidden state by a linear layer. 

The DGC module is extended for both speed, flow, and fluctuation terms. The structure of this extended DGC module is shown in the top figure of Fig.\ref{fig: modelarchi}. In this module, two different dynamic convolutional kernels are generated from the input and the hidden state for speed and flow in parallel. Another branch uses linear activated by the tanh function to give the node-wise fluctuations for speed and flow respectively. Here $R$ is the upper bound of fluctuations. Next, by using the dynamic kernels and the dynamic biases, graph convolution is applied and we get the predicted mean values of speed and flow. 

Directly training such an uncertainty-aware RNN-based traffic forecasting model is difficult because the uncertainty for each prediction step is unknown before training. To address this issue, the \emph{scheduled sampling} strategy \cite{bengio2015scheduled} is adapted here. The basic idea is that the probability of feeding the ground truth of the previous step to the decoder cells gradually decreases from 1 to 0 during training so the model starts from learning short-term predictions and then gradually learning long-term features. The same rule is also applied to uncertainty estimation. When the ground truth is given in the specific step of the decoder, the input uncertainty is fixed as a small value (\SI{0.2}{\kilo\meter\per\hour} in this study). It means that the total variance in Eq.\eqref{eq: total var} is set as 0.4. Therefore, the model focuses on quantifying one-step prediction uncertainty in the early stage of training and gradually learns how the prediction horizon impacts long-term uncertainty. In this study, we use the exponential function to control the probability of using ground truths:
\begin{equation}
    p = \exp{(-c \times i)}
\end{equation}

Here $i$ is the number of iterations and $c$ is a positive constant that controls the decay rate. In this study, we set $c = 1.25 \times 10^-4$ by trial and error. This strategy requires longer training time, which is another reason why we choose DEL--it is a single-pass method so it is not necessary to train many models or run many repeated inferences. Next, we will do a case study and investigate how much data is truly useful for data-driven traffic forecasting models.

\section{Experiments}
\label{sec: exp}

\begin{figure}[h!]
\centering
\includegraphics[width=0.45\linewidth]{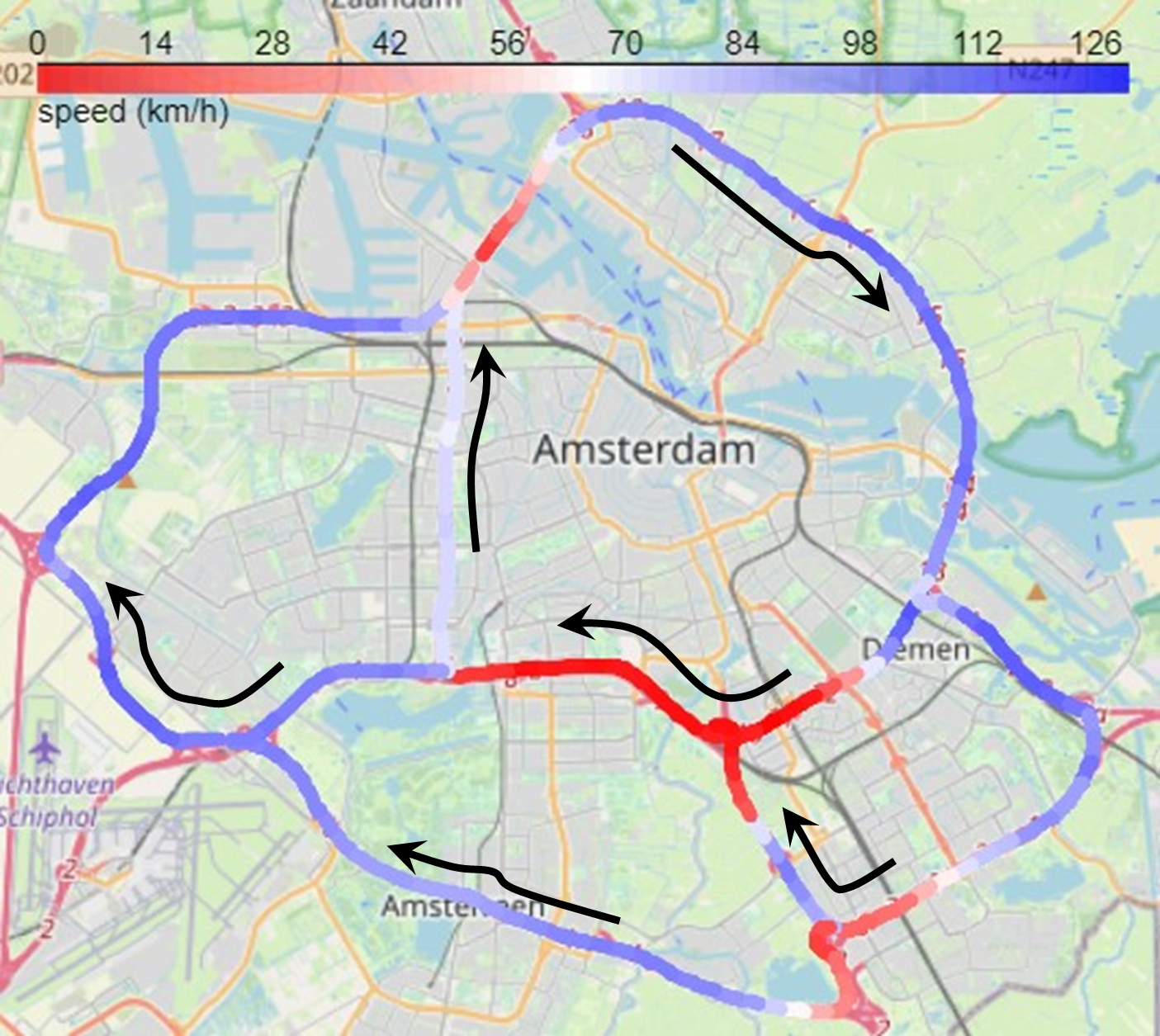}
\caption{The selected highway network. We only consider the clockwise driving directions in this study.} \label{fig: amsnet}
\end{figure}

In this section, we evaluate the effectiveness of the proposed method on a real-world highway network. The loop detector data utilized in this study is collected and processed by the National Data Warehouse (NDW - www.ndw.nu). Specifically, we focus on a heavily congested highway network in the Netherlands, which connects various suburban regions around Amsterdam and the Schipol International Airport. The highway network is depicted in Fig.\ref{fig: amsnet}. It is useful to note that we exclusively consider the clockwise driving directions for our analysis.

The data are aggregated every 2 minutes (time interval) and mapped on the highway in a uniform spatial resolution of \SI{400}{\meter} by the Adaptive Smoothing Method (ASM) \cite{treiber2003adaptive, schreiter2010two}. We prepared the speed and flow data during morning peak hours (7:00 - 10:00) and afternoon to evening period (15:00-21:00) from 1st January 2018 to 31st December 2022. The data set includes 5 years in total and only 14 days without data are excluded. The initial training set includes the years 2018 and 2019. Then the pre-trained model is deployed to give predictions and continuously collect data in 2020 and 2021. The data in 2022 is chosen as the test set. In this case study, the observation of speed and flow in the past \SI{40}{\min} is given (20 steps) and we aim to predict the traffic speed and flow in the next \SI{30}{\minute} (15 steps). 

To meet the condition in Eq.\eqref{eq: range boundary}, we set different graph convolution ranges (by adjusting the degree of adjacency matrices) for speed and flow branches in the extended DGC module. For speed prediction, we mainly forecast the congestion. The speed of the back-propagating stop-and-go wave is a stable value around \SI{0.3}{\kilo\meter\per\minute}. So the range of graph convolution should be between \SI{0.6}{\kilo\meter} and \SI{1.2}{\kilo\meter} (the time interval is \SI{2}{\minute}). Here we choose \SI{0.8}{\kilo\meter}, which means the degree of the adjacency matrix is 2 (2nd-order adjacent links are included in the graph convolution). For flow prediction, we must consider the forward propagating characteristic curves. The upper bound of the speed of this shock wave is approximately the speed limit, which is \SI{120}{\kilo\meter\per\hour}(\SI{2}{\kilo\meter\per\minute}). So the range of graph convolution should be between \SI{4}{\kilo\meter} and \SI{8}{\kilo\meter}. Here we choose \SI{4.4}{\kilo\meter}, which means the degree of the adjacency matrix for flow is 11.

More details of the model, hyperparameter setting, and an interactive web application that explains how the predictions are made can be found in the open-source code and the GitHub page \footnote{\url{https://github.com/RomainLITUD/uncertainty-aware-traffic-speed-flow-demand-prediction}}.

\section{Results}
\label{sec: results}

\subsection{Performance evaluation}
We first demonstrate that the prediction is reliable from the perspective of traffic flow theory. This evidence is important for validating that the proposed model indeed describes the desired traffic phenomena and thus the estimated uncertainty and the following results are trustworthy.

\begin{figure}[h!]
\centering
\includegraphics[width=0.7\linewidth]{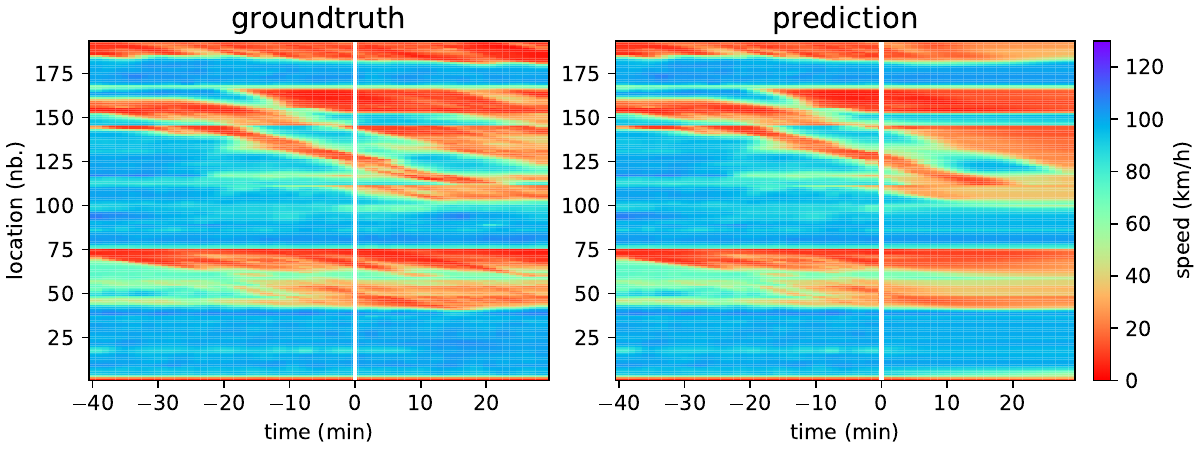}
\includegraphics[width=0.7\linewidth]{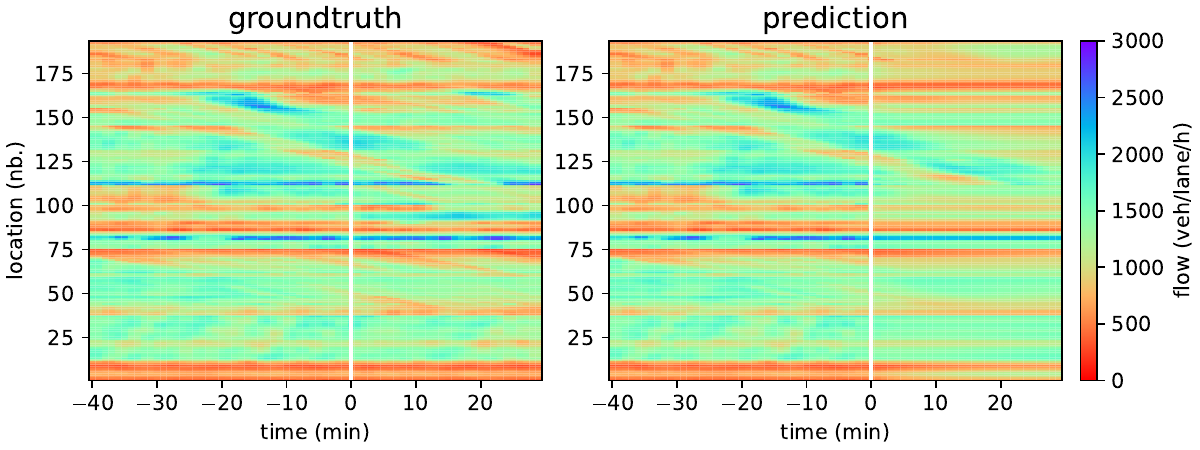}
\caption{Comparison of the speed/flow prediction and the ground truth. The vertical white lines mark the current timestamp. The y-axis is along the driving direction. The right side is the traffic state in the future.} \label{fig: prediction}
\end{figure}

Figure \ref{fig: prediction} shows an example of the predictions. The predictions are presented by stitching together different roads along their respective driving directions. It is clear that the model can forecast sharp and clear congestion patterns in the short-term future, including accurately matching spilling-back stop-and-go waves. However, as the prediction horizon extends to 10-15 minutes, the model's ability to provide precise contours of congestion patterns diminishes. The prediction becomes blurred, especially for predicted new congestion. This is primarily due to the increased uncertainty associated with long-term predictions. Long-term stop-and-go waves depend on incoming traffic demand and driving behaviours, which are inherently unpredictable when relying solely on loop detector data. So the mean value is between the congested and free-flowing states.
Nevertheless, the predictions intuitively reveal that the proposed model has successfully learned the correct spatiotemporal characteristics of traffic flow. This evidence validates the ability of the proposed model to describe the desired traffic phenomena, thereby affirming the trustworthiness of the estimated uncertainty and subsequent results.

\begin{figure}[h!]
\centering
\includegraphics[width=0.5\linewidth]{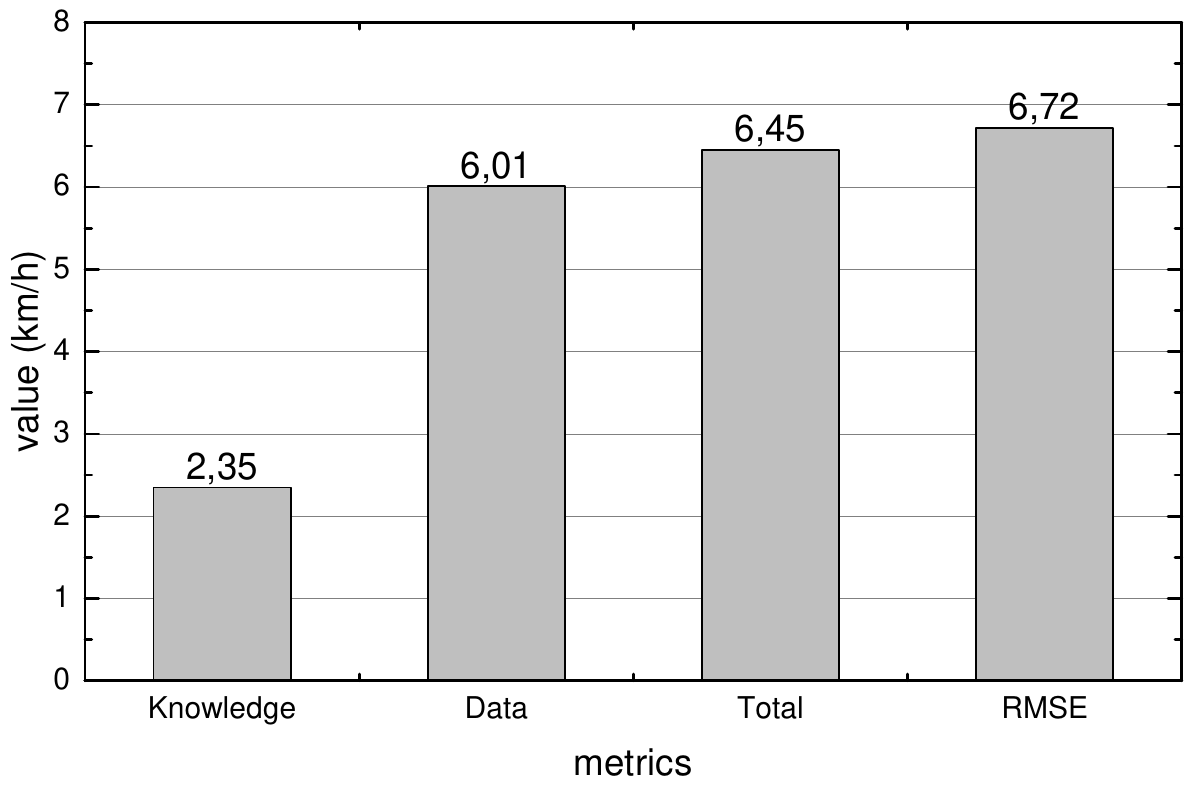}
\caption{The values of different uncertainty components and the measured RMSE of speed on the test set.} \label{fig: calibrate}
\end{figure}

Next, we check whether the estimated uncertainty is consistent with the prediction accuracy of speed. Fig.\ref{fig: calibrate} shows the mean squared values of data, knowledge, and total uncertainty of speed prediction on the test set. They are also compared with the measured RMSE of speed. The total uncertainty is indeed close to RMSE. The difference is less than \SI{0.3}{\kilo\meter\per\hour}. This consistency also supports that the proposed model precisely estimates the uncertainty. We also observe that the average data uncertainty is significantly higher than the average knowledge uncertainty. It means that the inherent unpredictable part is the main factor that restricts the model's performance. The lower bound of RMSE for speed prediction is around 6 km/h, suggesting that there are no well-formulated models (not over-fitted to biases) that can outperform this value.

\subsection{Training set distillation}
In this subsection, we investigate the informativeness of the training set for traffic forecasting. We first rank all samples in the training set in ascending order based on their average estimated knowledge uncertainty of 15-step predictions. Next, we selectively preserve or remove samples based on their knowledge uncertainty percentile. For instance, if we want to preserve the lowest X\% of samples, we retain those samples whose knowledge uncertainty falls within this range and remove the rest. After creating the new data set by preserving or removing samples, we re-train the model using the new data set. Finally, we assess how the accuracy of the same model changes with different percentages of preserved and removed samples. This analysis allows us to determine which portion of the training data contains the most informative samples for traffic speed forecasting.

\begin{figure}[h!]
\centering
\includegraphics[width=0.49\linewidth]{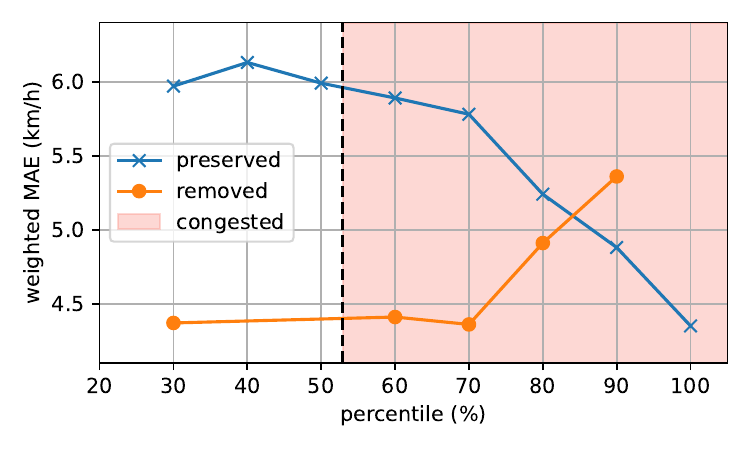}
\includegraphics[width=0.49\linewidth]{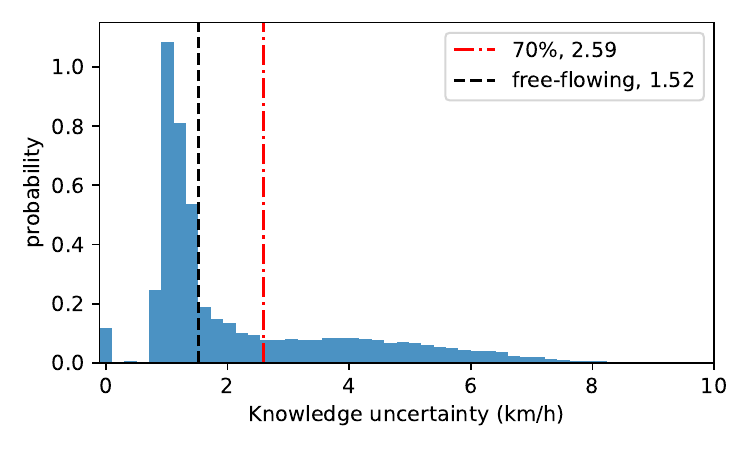}
\caption{(a) Relationships between the weighted MAE and the removed/preserved percentage; (b) The distribution of knowledge uncertainty of the training set.} \label{fig: distill}
\end{figure}

To better show the quality of congestion prediction, we use the weighted MAE of speed here. For each predicted speed, if the ground truth is below \SI{60}{\kilo\meter\per\hour} (congested), the MAE is multiplied by 4, otherwise the coefficient is 1. So, the contribution of congested areas and periods is inflated. The result is presented in Fig.\ref{fig: distill}. The vertical dotted line separates the free-flowing samples (left blank area) from the samples containing congestion (right pink area). This is reasonable because free-flowing traffic is highly predictable. When we ``preserve'' the lowest x\% samples, with the increasing of x, the prediction accuracy keeps decreasing in the congested area and the slope becomes steeper at the high knowledge uncertainty end. The tendency of this curve supports that high knowledge uncertainty samples contribute more to model training than low knowledge uncertainty samples. When we ``remove'' the lowest x\% samples, with the increasing of x, the accuracy keeps horizontal in a wide range but suddenly increases at 70\%. The result reveals that a substantial portion of the training set, approximately at 70\%, does not significantly contribute to the prediction power of trained models. Removing these samples has little to no influence on the model's performance of speed prediction. This observation is crucial as it indicates that we can distil the training set by discarding approximately 70\% of the samples. Consequently, we can retain only the top 30\% of samples characterized by higher knowledge uncertainty.

Fig.\ref{fig: distill}(b) further gives the distribution of average knowledge uncertainty in the training set. The vertical black line divides the traffic states into congested and free-flowing categories. Free-flowing samples are at the lower end of the knowledge uncertainty spectrum. Samples between the black and red lines represent periods containing only local and non-connected congestion patterns. The 70 percentile value is recorded at \SI{2.59}{\kilo\meter\per\hour}. On the right side of this threshold, heavily congested periods are observed, and they are considered informative and valuable for training traffic forecasting models. This finding suggests that the features of local congestion patterns are included in those large-scale congestion patterns. To filter the data stream from 2020 and 2021, we will set the threshold to this value.

\subsection{Online data collection}
Now, we mimic the real-time data stream collection process. The pre-trained model is applied to the data set of 2020 and 2021 for identifying those samples whose knowledge uncertainty is higher than the threshold (\SI{2.59}{\kilo\meter\per\hour}). 13.4\% samples in 2020 and 16.1\% samples in 2021 fall into this range. The ratio is lower than in 2018 and 2019 because of the Covid measures in the Netherlands. These samples are combined with the top 30\% samples of the training set to compose the new ``distilled'' training set. We re-trained the model again on this distilled data set and compared its performance with (1) the model pre-trained on the initial training set and (2) another model trained on the initial training set plus the un-selected samples in 2020 and 2021. The sizes and the performances of the 3 models on the test set are compared in Table.\ref{tab: performance compare}. 

\begin{table}[h!]
	\caption{Prediction accuracy of the same model trained on different training sets}\label{tab: performance compare}
	\centering
		\begin{tabular}{lcccc}
	        \hline    
            \multicolumn{5}{c}{speed}\\
            \hline
            Training set & MAE(\SI{}{\kilo\meter\per\hour}) & MAPE(\%) & RMSE(\SI{}{\kilo\meter\per\hour}) & Nb of samples\\
            \hline
             Initial training set & 3.01 & 4.98 & 6.99 & 34215\\
             Distilled set & \textbf{2.90} & \textbf{4.88} & \textbf{6.70} & 15568\\
             Initial + unselected & 3.01 & 5.04 & 6.82 & 70345\\
             \hline
             \multicolumn{5}{c}{flow}\\
            \hline
            Training set & MAE(veh/lane/h) & MAPE(\%) & RMSE(veh/lane/h) & Nb of samples\\
             \hline
             Initial training set & 118.8 & 14.86 & 171.6 & 34215\\
             Distilled set & \textbf{110.6} & \textbf{13.60} & \textbf{155.28} & 15568\\
             Initial + unselected & 116.0 & 14.11 & 166.9 & 70345\\
             \hline
        \end{tabular}
\end{table}

The results demonstrate that the distilled training set, despite having significantly fewer samples, leads to the best model performance, especially for speed prediction, due to the high informativeness of the retained samples. The smaller size and informative samples also contribute to shorter training times and faster convergence in training. In contrast, although the unselected set is significantly larger, yet we observe a smaller improvement in accuracy. This result suggests that the majority of loop detector data may lack relevance and validity for traffic forecasting. What proves crucial for training traffic forecasting models is the presence of truly "informative" data rather than simply a larger data set. However, it is important to acknowledge that even with the informative distilled training set, the improvement in model performance remains marginal for speed prediction. This marginal improvement is unsurprising and attributed to the inherent high data uncertainty present in traffic dynamics—the unpredictable component of traffic patterns, as illustrated in Fig.\ref{fig: calibrate}.

Additionally, it is useful to recall that the data set distillation is based on the uncertainty of speed prediction. If the same process is applied to traffic flow, the result might be different.

\section{Conclusion and perspective}
\label{sec: conc}

This study focuses on quantifying the information content of data for traffic prediction. 
Do (e.g. loop detector) data sets provide sufficient and diverse information to support training large and complex data-driven prediction models? 
To this end, we propose a unified traffic forecasting framework that can give reliable predictions and estimates of the corresponding uncertainty. 
The experiments and the results reveal
that a large proportion of the macroscopic highway loop-detector data are non-informative (free-flowing and local recurring congestion patterns). The case study of a highway network around Amsterdam shows that among the 4-year data (2018-2021), only around 18\% of the peak hour data needs to be preserved. Removing the rest of the data hardly influences the performance of data-driven models trained on this data set. 

Since in our case, a single data sample encompasses all loop detectors of the entire network (for a single time period), these condensed data sets may be reduced even more removing data that are ``spatially redundant''. This could pertain, for example, to data from closely spaced loop detectors on stretches with no on and off ramps between the sensors. Adding this additional ``data distillation'' process could be done through global sensitivity analyses and/or model pruning, we plan on extending our method in this direction in the future.

Our results also give rise to a more tentative reflection on paths to improve the predictive performance of the models.
We argue that a key issue in the traffic forecasting domain is that there is a surplus of modelling techniques (particularly in the ML domain) but an insufficient amount of effective data sets (in terms of information content) to train these models in ways that lead to methods that really can be generalized. This is on the one hand due to limitations in the observability of many of the governing variables that result in traffic in the first place (e.g. demand, route choice); and on the other hand due to the large heterogeneity in circumstances (e.g. different networks, control and many other influence factors). 
The question then becomes how to fuse sufficiently heterogeneous data from multiple sources (e.g. trajectories, demand, or even social media data) to ``distil'' sufficient amounts of information into a training data set so that generally applicable prediction methods may evolve. In line with this question, how much heterogeneity in the used data sets is needed to reduce the prediction bounds for a given case? Testing with multiple large and heterogeneous data sets from different cities is necessary to explore these questions.

\section{Acknowledgements}
This research is sponsored by the NWO/TTW project MiRRORS with grant agreement number 16270. We thank them for supporting this study.

\bibliographystyle{unsrt}  
\bibliography{references}  


\end{document}